\documentclass{article}

 \usepackage[preprint]{neurips_2025}

\usepackage[utf8]{inputenc} 
\usepackage[T1]{fontenc}    
\usepackage{hyperref}       
\usepackage{url}            
\usepackage{booktabs}       
\usepackage{amsfonts}       
\usepackage{nicefrac}       
\usepackage{microtype}      
\usepackage{xcolor}         
\usepackage{graphicx}       
\usepackage{soul}
\usepackage{enumitem}
 \usepackage{amsmath}
 \usepackage{bbm}
\workshoptitle{New Perspectives in Advancing Graph Machine Learning}
\title{Spatio-Temporal Directed Graph Learning for Account Takeover Fraud Detection}

\author{%
  Mohsen Nayebi Kerdabadi \\
  \texttt{mohsen.nayebikerdabadi@capitalone.com}
  \AND
  William Andrew Byron \\
  \texttt{drew.byron@capitalone.com}
  \AND
  Xin Sun \\
  \texttt{xin.sun@capitalone.com}
  \AND
  Amirfarrokh Iranitalab\thanks{Corresponding author.} \\
  \texttt{amirfarrokh.iranitalab@capitalone.com}
  \vspace{1.5em} \\
  AI Foundations, Capital One, McLean, Virginia, United States
}

\begin{document}

\maketitle

\vspace{-5mm}
\begin{abstract}
\vspace{-3mm}
Account Takeover (ATO) fraud poses a significant challenge in consumer banking, requiring high recall under strict latency while minimizing friction for legitimate users. Production systems typically rely on tabular gradient-boosted decision trees (e.g., XGBoost) that score sessions independently, overlooking the relational and temporal structure of online activity that characterizes coordinated attacks and “fraud rings.” We introduce \textbf{ATLAS} (\textbf{A}ccount \textbf{T}akeover \textbf{L}earning \textbf{A}cross \textbf{S}patio-Temporal Directed Graph), a framework that reformulates ATO detection as spatio-temporal node classification on a time-respecting directed session graph. ATLAS links entities via shared identifiers (account, device, IP) and regulates connectivity with time-window and recency constraints, enabling causal, time-respecting message passing and latency–aware label propagation that uses only labels available at scoring time, non-anticipative and leakage-free. We operationalize ATLAS with inductive GraphSAGE variants trained via neighbor sampling, at scale on a sessions graph with 100M+ nodes and \textasciitilde{}1B edges. On a high-risk digital product at Capital One, ATLAS delivers +6.38\% AUC and $>\!50\%$ reduction in customer friction, improving fraud capture while reducing user friction.
\end{abstract}

\vspace{-4mm}
\section{Introduction}

Account Takeover (ATO) fraud occurs when an adversary gains unauthorized access to a legitimate customer account via credential stuffing, phishing, device spoofing, or related tactics, and initiates high-risk transactions (HRTs). HRTs are monetizable actions (e.g., funds transfers) that are prime targets for fraudsters. In consumer banking, ATO drives both direct financial loss and customer-experience degradation. Therefore, effective defenses must increase fraud capture while minimizing friction for legitimate users.

A naïve approach is to impose friction on every HRT in every online session (e.g., additional verification). While simple, blanket friction creates a substantial customer burden, raising abandonment and complaints, and can erode trust and retention. We instead adopt a risk-based approach: train a model to produce a real-time risk score per HRT online session and apply additional friction only to the high-risk subset. This reduces overall friction while maintaining high fraud recall, yielding a more favorable capture–friction trade-off.

Practically, ATO detection in banking faces extreme class imbalance, rapidly evolving attacker behavior (concept drift), and a strict online latency budget (e.g., $<\!250$\,ms). In this setting, production systems have long relied on tabular gradient-boosted decision trees, XGBoost \citep{chen2016xgboost}, that score each session independently from engineered numerical features. Despite extensive trials with deep architectures (fully connected neural networks \citep{goodfellow2016deep}, RNNs \citep{hochreiter1997long}, Transformers \citep{vaswani2017attention}), none consistently surpassed the boosted baseline under comparable latency and reliability constraints. Consequently, most gains have historically come from feature-engineering improvements atop XGBoost, which remains the dominant production solution.

Although XGBoost has established itself as the dominant in-production solution for tabular data \citep{shwartz2022tabular}, it scores each session in isolation, effectively assuming independent and identically distributed (i.i.d.) observations. This flat, per-row view ignores the relational structure (entity linkages via account/device/IP) and temporal structure (causal ordering/recency) that characterize coordinated campaigns (“fraud rings”). As a result, the model cannot transfer risk across connected sessions and discards high-signal cues such as historical neighbor labels and temporal dependencies. Crucially, these shortcomings are intrinsic to the tabular formulation: collapsing sessions into independent rows prevents any per-session learner (XGBoost, MLP, or per-row Transformer) from representing time-respecting edges, path-based evidence sharing, or non-anticipative (serve-time) past fraud label availability; capturing these signals requires an explicit spatio-temporal structure representation.

To close this gap, we introduce \textbf{ATLAS} (\textbf{A}ccount \textbf{T}akeover \textbf{L}earning \textbf{A}cross \textbf{S}patio-Temporal Directed Graph), which reformulates ATO detection as spatio-temporal node classification on a time-respecting directed session graph. This framing enables time-respecting message passing across connected sessions and lag-aware (partially observed) label propagation. More specifically, our contributions are:
\begin{itemize}[leftmargin=*]
    \item We reformulate ATO as spatio-temporal graph learning with node classification over sessions on a time-respecting directed acyclic graph, enabling causal, time-respecting message passing and lag-aware label propagation that incorporates only serve-time-available historical evidence.
    \item We construct a directed temporal graph with strict causal ordering (past-to-future); link entities via shared identifiers (account, device, IP); and regulate connectivity with designed time-window and recency constraints. We operationalize this with an inductive GraphSAGE \citet{hamilton2017inductive} encoder trained via neighbor sampling, ensuring serve-time consistency and latency compliance.
    \item We conduct extensive experiments on a high-risk digital product dataset at Capital One financial institution (100M+ nodes, \textasciitilde{}1B edges), demonstrating the effectiveness of our approach: +6.38\% AUC and $>\!50\%$ reduction in customer friction, improving fraud capture while lowering friction.
\end{itemize}

\section{Method}

We present ATLAS, a spatio-temporal directed-graph approach to ATO with three parts: (i) a time-respecting session graph with causal edges and connectivity controlled by window $T$ and recency cap $K$ (Section ~\ref{gfs}), (ii) serve-time–consistent (non-anticipative) label aggregation that appends lagged neighbor-label features (Section ~\ref{sec:labelprop}), and (iii) an inductive GraphSAGE encoder with mini-batch neighbor sampling and relational/time-aware/attention variants (Section ~\ref{sec:gnn}). This yields leakage-free features and latency-compliant inference.

\subsection{Graph Formulation Strategy}
\label{gfs}

\begin{figure}[t]
    \centering
    \includegraphics[width=0.82\linewidth]{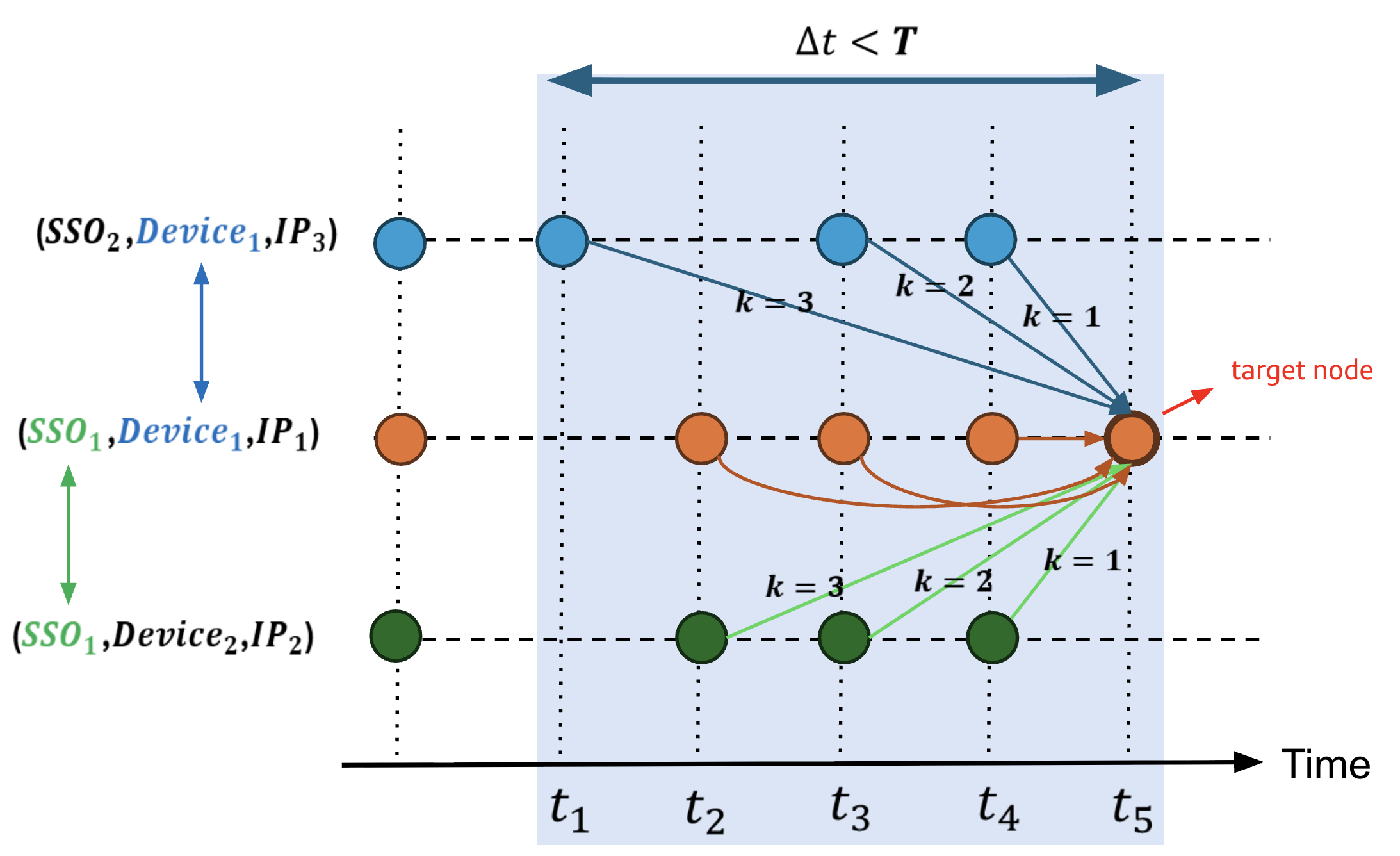}
    \caption{\textbf{ATLAS graph formulation.} Nodes are HRT sessions keyed by
    (account\_id, device\_id, ip\_address, timestamp). Directed edges point past$\!\to\!$future between sessions sharing an identifier, restricted by time window \(T\) and per-identifier recency cap \(K\). Edge types correspond to the linking identifier (account/device/IP).}
    \label{figure1}
\end{figure}

We model ATO as spatio-temporal node classification on a directed session graph \(G=(V,E)\) depicted in Figure ~\ref{figure1}. Each node \(v\in V\) is a high-risk transaction (HRT) session scored at serve time; edges \(e\in E\) encode temporal, identifier-based links from prior sessions.

\textbf{Nodes.} Each node \(v\in V\) is uniquely keyed by (account\_id, device\_id, ip\_address, timestamp) and carries a feature vector \(\mathbf{x}_v \in \mathbbm{R}^d\) (curated tabular features) and a binary label \(y_v\in\{0,1\}\) (1 = fraud, 0 = non-fraud). The model outputs a risk score
\(
s_v \;=\; \Pr\!\big(y_v=1 \mid G_{\prec v},\, X_{\preceq v}\big)\in[0,1],
\)
where \(G_{\prec v}\) is the subgraph of past neighbors of \(v\) (strictly earlier sessions) and
\(
X_{\preceq v}=\{\mathbf{x}_u:\, u\in V,\ t_u\le t_v\},
\)
ensuring non-anticipative, leakage-free scoring.

\textbf{Edges.}
For sessions \(u\) and \(v\) with timestamps \(t_u<t_v\), we add a directed edge
\((u\!\to\!v)\) if they share an identifier
\(m\in\mathcal{M}=\{\text{account\_id}, \text{device\_id}, \text{ip\_address}\}\).
Edges are typed by \(m\), yielding \(E=\bigcup_{m\in\mathcal{M}} E_m\). Because \(t_u<t_v\), the graph is time-respecting (acyclic).

\textbf{Graph Connectivity Regulation.}
To ensure non-anticipative connectivity and control degree/latency, we enforce two connectivity constraints on graph edges :
\begin{itemize}[leftmargin=*]
  \item \emph{Time-window} \(T\): include \((u\!\to\!v)\) only if \(0 < t_v-t_u \le T\).
  \item \emph{Recency cap} \(K\): for each \(v\) and type \(m\), keep at most the \(K\) most recent predecessors in \(E_m\).
\end{itemize}
These constraints (i) preserve causal ordering, (ii) prioritize informative recent history, and (iii) cap neighborhood size for stable sampling and serve-time latency.

\textbf{Resulting Structure.}
The graph exposes cross-session patterns (e.g., coordinated “fraud rings”) and supports causal, time-respecting message passing and serve-time–consistent lagged-label features (Section ~\ref{sec:labelprop}), as well different GraphSAGE variants with neighbor sampling at scale (Section .~\ref{sec:gnn}).

\subsection{Lag-aware Label Propagation}
\label{sec:labelprop}

To incorporate historical evidence without leakage, we augment each session node with labels from past, connected sessions whose ground truth is already known at the target’s serve time. Let \(v\) denote the target session with serve time \(t_v\). For any earlier session \(u\) with timestamp \(t_u<t_v\), the binary label \(y_u\in\{0,1\}\) becomes available at adjudication time \(\tau_u\). We therefore restrict propagation to labels that would be known at serve time by enforcing \(\tau_u \le t_v\).

Starting from the directed session graph \(G=(V,E)\) (Section~\ref{gfs}), we collect the in-neighborhood of \(v\) within a finite history window \(T>0\):
\begin{equation}
\mathcal{N}^{-}_{T}(v) \;=\; \bigl\{\, u \;:\; (u\!\to\!v)\in E,\; 0 < t_v - t_u \le T \,\bigr\}.
\end{equation}
To control degree and latency, we keep only the \(K\) most recent predecessors:
\begin{equation}
\mathcal{R}(v) \;=\; \operatorname*{TopK}_{u\in \mathcal{N}^{-}_{T}(v)}\!\bigl(t_u\bigr),
\end{equation}
so that \(|\mathcal{R}(v)|\le K\). We then apply the delayed-label filter to mirror what is available online:
\begin{equation}
\mathcal{A}(v) \;=\; \bigl\{\, u\in \mathcal{R}(v) \;:\; \tau_u \le t_v \,\bigr\}.
\end{equation}
Here, \(T\) (e.g.,~days) bounds how far back we look, \(K\) caps neighborhood size for stable sampling and low latency, and the condition \(\tau_u\le t_v\) ensures causal, serve-time–correct neighbor-label features.

From the available set \(\mathcal{A}(v)\) we compute simple aggregates:
\begin{align}
n^{\mathrm{lab}}_{v}   &= \bigl|\mathcal{A}(v)\bigr|                       && \text{(count of neighbors with known labels)}\\
n^{\mathrm{fraud}}_{v} &= \sum_{u\in \mathcal{A}(v)} y_u                   && \text{(count of known-fraud neighbors)}\\
r_{v}                  &= \frac{n^{\mathrm{fraud}}_{v}}{\max\!\bigl(1,\,n^{\mathrm{lab}}_{v}\bigr)} && \text{(empirical fraud rate among known labels)}\\
a_{v}                  &= \mathbbm{1}\!\left[n^{\text{fraud}}_{v}\ge 1\right]       && \text{(any known fraud upstream)}
\end{align}

We then form a label-propagation feature vector
\begin{equation}
\ell_v \;=\; \bigl[\, n^{\mathrm{lab}}_{v},\; n^{\mathrm{fraud}}_{v},\; r_v,\; a_v \,\bigr]
\end{equation}
and append it to the node input used by the encoder:
\begin{equation}
\mathbf{h}^{(0)}_v \;=\; \bigl[\, \mathbf{x}_v \,;\, \ell_v \,\bigr],
\end{equation}
where \(\mathbf{x}_v\) are the curated tabular features. During training, we apply the same \(T\), \(K\), and \(\tau_u\le t_v\) rules to avoid training–serving skew. If no labels are available (\(n^{\mathrm{lab}}_{v}=0\)), we set \(r_v=0\) and \(a_v=0\). This design preserves strict causal ordering, aligns with delayed supervision in production, and provides the model with lightweight, high-signal context from truly known historical outcomes.

\subsection{GNN Architecture}
\label{sec:gnn}

Our encoder is based on GraphSAGE \citep{hamilton2017inductive}, chosen for its inductive capability and support for mini-batch neighbor sampling, enabling training at our large scale. Each node begins with
$\mathbf{h}^{(0)}_v \;=\; \bigl[\, \mathbf{x}_v \,;\, \ell_v \,\bigr]$
where \(\mathbf{x}_v\) are curated tabular features and \(\ell_v\) are lagged label features (Section ~\ref{sec:labelprop}).

\paragraph{Homogeneous GraphSAGE.}
We construct a time-respecting \(h\)-hop ego-graph around \(v\) using a neighbor sampler that enforces the same \(T\) and \(K\) constraints (Section~\ref{gfs}). Let the per-hop fanouts be \(\mathbf{f}=(f_1,\ldots,f_h)\) (at most \(f_k\) past neighbors sampled at hop \(k\); when constant, \(f_k\equiv f\) for all \(k\)). For layer \(k=1,\ldots,L\) (with \(L\le h\)), let \(\mathcal{S}^{(k)}(v)\subseteq \mathcal{N}^{-}_{T}(v)\) denote the sampled in-neighbors used at layer \(k\). A GraphSAGE block aggregates neighbor states and updates the target:
\begin{align}
\mathbf{m}_v^{(k)} &= \operatorname{AGG}^{(k)}\!\Bigl(\bigl\{\mathbf{h}^{(k-1)}_u : u\in\mathcal{S}^{(k)}(v)\bigr\}\Bigr),\\
\mathbf{h}^{(k)}_v &= \sigma\!\Bigl(W^{(k)}\bigl[\mathbf{h}^{(k-1)}_v \,;\, \mathbf{m}_v^{(k)}\bigr] + \mathbf{b}^{(k)}\Bigr),\label{eq:sage-update}
\end{align}
optionally followed by \(\ell_{2}\)-normalization and dropout. We use \emph{mean} as the aggregator \(\operatorname{AGG}^{(k)}\).

\paragraph{Relational GraphSAGE.}
As edges are typed by identifier (account/device/IP), we use a relational variant that aggregates per type and then fuses the results. Let \(\mathcal{M}\) be the set of relation types and let
\(\mathcal{S}^{(k)}_{m}(v) \subseteq \mathcal{N}^{-}_{T,m}(v)\) denote the sampled in-neighbors of type \(m\) used at layer \(k\). We compute
\begin{equation}
\mathbf{m}_{v}^{(k)}
\;=\;
\sum_{m\in\mathcal{M}}
\Phi^{(k)}_{m}\!\left(
  \operatorname{AGG}^{(k)}_{m}
  \Bigl(\{\mathbf{h}^{(k-1)}_u \;:\; u\in \mathcal{S}^{(k)}_{m}(v)\}\Bigr)
\right),
\end{equation}
where \(\operatorname{AGG}^{(k)}_{m}\) is a type-specific aggregator (we use \emph{mean}) and
\(\Phi^{(k)}_{m}\) is a learnable type-specific transform (e.g., a linear map or gate). The node update then follows equation \ref{eq:sage-update}.

\paragraph{Time-aware / attention variant.}
To encode recency and relation importance, we incorporate simple edge features
(e.g., binned \(\Delta t=t_v-t_u\) and edge-type embeddings) either by concatenation
into the neighbor vector or via an attention aggregator:
\begin{align}
\alpha^{(k)}_{vu}
&= \operatorname{softmax}_{\,u\in \mathcal{S}^{(k)}(v)}
  \Bigl(\mathbf{a}^{\top}\bigl[\, W_{\mathrm{qry}}\mathbf{h}^{(k-1)}_v \,;\, W_{\mathrm{key}}\mathbf{h}^{(k-1)}_u \,;\, \mathbf{e}_{uv}\,\bigr]\Bigr), \label{eq:attn-weights}\\
\mathbf{m}_v^{(k)}
&= \sum_{u\in \mathcal{S}^{(k)}(v)} \alpha^{(k)}_{vu}\, W_{\mathrm{val}}\mathbf{h}^{(k-1)}_u, \label{eq:attn-agg}
\end{align}
where \(\mathcal{S}^{(k)}(v)\subseteq \mathcal{N}^{-}_{T}(v)\) is the sampled in-neighborhood at layer \(k\),
\(\mathbf{e}_{uv}\) encodes the (binned) time gap and relation type for edge \((u\!\to\! v)\),
and \(W_{\mathrm{qry}}, W_{\mathrm{key}}, W_{\mathrm{val}}\) and \(\mathbf{a}\) are learnable parameters.
Optionally, a multi-head variant can replace \eqref{eq:attn-weights}–\eqref{eq:attn-agg} with
head-wise projections and concatenation.

\paragraph{Neighbor sampling and depth.}
We train with mini-batches of seed nodes and per-layer fanouts \((f_1,\ldots,f_L)\), sampling \(\mathcal{S}^{(k)}(v)\subseteq \mathcal{N}^{-}_{T}(v)\) under the same \((T,K)\) constraints used at serve time to avoid train–serve skew. In practice, shallow depth \((L\in\{2,3\})\) with moderate fanouts provides a good accuracy–latency trade-off.

\paragraph{Output and loss.}
The final embedding \(\mathbf{h}^{(L)}_v\) is passed to a logistic head
\begin{equation}
s_v=\sigma(\mathbf{w}^{\top}\mathbf{h}^{(L)}_v + b).
\end{equation}
We optimize a weighted binary cross-entropy to address class imbalance. Decision thresholds are calibrated to the target friction envelope.

\section{Results}
\label{results}

We compare our GNNs to the production XGBoost baseline and ablate graph hyperparameters. Beyond a simple homogeneous GraphSAGE, added architectural complexity yields little benefit; most gains come from the graph formulation and serve-time–consistent lagged labels. Prior trials with FNNs and tabular Transformers matched XGBoost under the same features/latency, reinforcing that improvements stem from exploiting graph structure rather than deeper per-row models.

\subsection{Dataset} The dataset comprises tens of millions of sessions with a very low ATO base rate (extreme class imbalance). To evaluate generalization to future traffic, we use a chronological split: 8 months for training, 2 months for validation, and 5 months for testing (no overlap). All numerical features are standardized using statistics computed on the training set only. We assemble features, labels, and edge indices into PyTorch Geometric (PyG) data objects and employ PyG's NeighborLoader for efficient, out-of-core neighborhood sampling. This dynamic loading is essential for a continuously growing graph. Owing to data sensitivity and confidentiality, we do not report descriptive statistics. The corpus contains two major segments from a digital product at Capital One; due to confidentiality, we anonymize them as \textit{Segment 1} and \textit{Segment 2}.

\subsection{Performance Comparison}

The GNN consistently outperforms XGBoost for both segments. As shown in Table~\ref{table1}, the GNN achieves an overall ROC AUC of 82.27 (vs.\ 79.83 for XGBoost), a \emph{+3.06\%} relative improvement. By segment, gains are \emph{+3.43\%} (Segment~1) and \emph{+1.66\%} (Segment~2). The strongest results come from homogeneous GraphSAGE \emph{with} label propagation, yielding an overall ROC AUC of 84.46 and a \emph{+5.8\%} relative improvement over XGBoost.

\begin{table}
  \caption{Performance comparison of XGBoost (XGB) vs.\ GNN, with and without label propagation. (ROC AUC reported as percentages; improvements are relative to XGB.)}
  \small
  \label{table1}
  \centering
  \begin{tabular}{lccc}
    \toprule
    \textbf{Model} & \textbf{AUC Overall} & \textbf{AUC Segment 1} & \textbf{AUC Segment 2} \\
    \midrule
    XGB & 79.83 & 78.88 & 82.45 \\
    \midrule
    GNN & 82.27 & 81.59 & 83.82 \\
    \textit{Improvement (vs.\ XGB)} & \textit{+3.06\%} & \textit{+3.43\%} & \textit{+1.66\%} \\
    \midrule
    GNN + Label Propagation & 84.46 & 83.92 & 85.45 \\
    \textit{Improvement (vs.\ XGB)} & \textit{+5.8\%} & \textit{+6.38\%} & \textit{+3.63\%} \\
    \bottomrule
  \end{tabular}
\end{table}

\subsection{Hyperparameter Analysis (K and T)}

\begin{figure}
    \centering
    \includegraphics[width=1.00\linewidth]{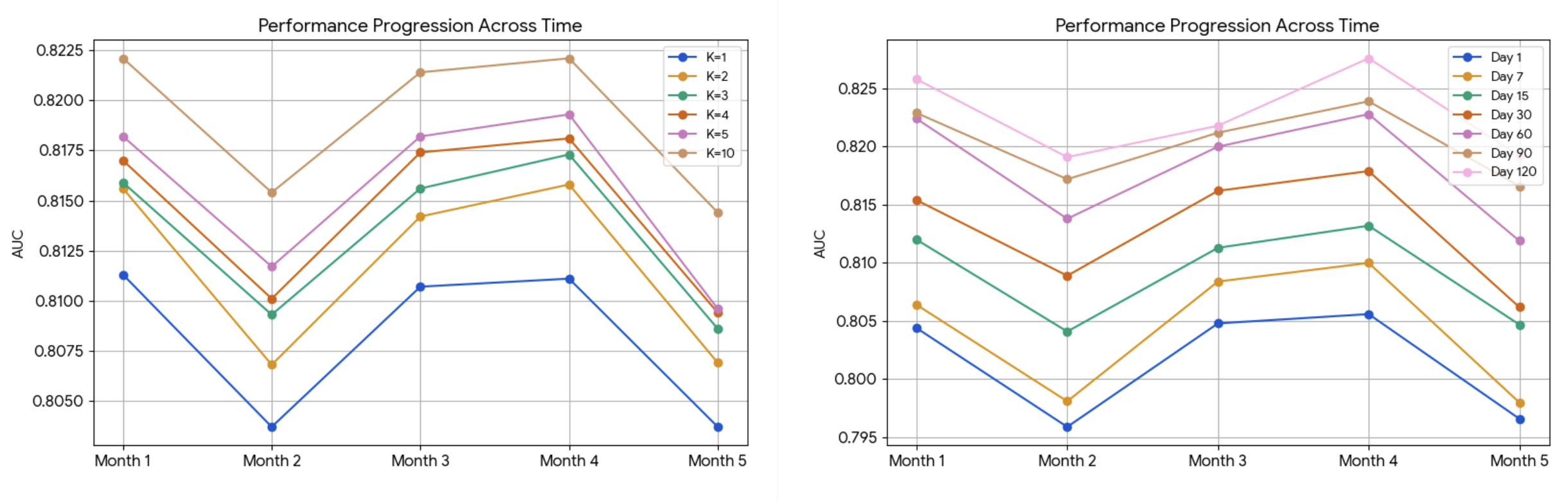}
    \caption{Effect of \(K\) (left) and \(T\) (right) on ROC AUC for Segment~1; similar trends hold for Segment~2.}
    \label{figure2}
\end{figure}

We study the recency cap \(K\) and time window \(T\). As illustrated in Figure~\ref{figure2}, increasing \(K\) from 1 to 10 steadily improves ROC AUC, indicating that incorporating more \emph{recent} historical sessions is beneficial. Likewise, extending \(T\) from 1 to 120 days yields consistent gains, underscoring the value of a longer temporal context for detecting coordinated activity.

\section{Related Work}
\vspace{-1mm}
ATO detection in industry commonly relies on tabular gradient-boosted trees such as XGBoost \citep{chen2016xgboost}; tree ensembles often remain strong on structured data \citep{shwartz2022tabular}, but they struggle to capture cross-session relational patterns.

Foundational GNNs include GCNs for node classification \citep{kipf2017semisupervised}, inductive GraphSAGE with neighbor sampling \citep{hamilton2017inductive}, and attention-based GAT \citep{velickovic2018graph, kerdabadi2025multi, hadizadeh2025discovering}. For evolving interactions, temporal models such as Temporal Graph Networks (TGN) \citep{rossi2020temporal}, TGAT \citep{xu2020inductive}, and DySAT \citep{sankar2020dysat} incorporate time into message passing. Our work differs by enforcing a time-respecting DAG and serve-time constraints to achieve non-anticipative inference under strict latency at enterprise scale.

Recent graph-based approaches to transactional fraud explicitly model accounts, devices, and transactions as graphs with temporal or entity-sharing edges. Semi-supervised credit-card fraud detection via attribute-driven graph representations treats users/transactions as nodes and propagates attribute signals \citep{xiang2023semi}. For edge-level scoring, FraudGT applies a graph transformer with edge-aware attention and message gating \citep{lin2024fraudgt}. Under sparse labels, Barely Supervised learning introduces structure-aware contrastive objectives \citep{yu2024barely}. On heterogeneous graphs, DGA-GNN combats noisy neighborhoods via dynamic grouping \citep{duan2024dga}. Low-homophily settings motivate label-aware aggregation and transformer encoders \citep{wang2023label}. 

\section{Conclusion}
\vspace{-1mm}

We presented ATLAS, a spatio-temporal graph framework for ATO detection that operates on a time-respecting directed session graph with connectivity regulated by a time window and recency cap. By combining serve-time-consistent lagged label aggregation with inductive GraphSAGE variants and neighbor sampling, ATLAS scales to 100M+ nodes and \textasciitilde1B edges while remaining latency compliant. On a high-risk digital product, it yields +6.38\% AUC and $>\!50\%$ reduction in customer friction, improving fraud capture and user experience. Due to privacy and regulatory constraints at Capital One, we cannot release data or detailed dataset statistics, and our evaluation is limited to anonymized segments of an online digital product. 

\bibliographystyle{abbrvnat}
\bibliography{references}
\newpage

\end{document}